\documentclass[preprint,12pt]{elsarticle}

\usepackage{amssymb}

\journal{Engineering Applications of Artificial Intelligence}

\usepackage{caption}

\def\state{s}
\def\stateSpace{\mathbb{S}}
\def\ctrl{a}
\def\ctrlSpace{\mathbb{A}}

\def\real{\mathbb R}

\def\est{\widehat}

\def\ord#1{\langle #1 \rangle}
\def\Buy{\text{Buy}}
\def\Sell{\text{Sell}}
\def\normal{\mathcal{N}}

\def\comment#1{}

\def\eqref#1{(\ref{#1})}
\def\Beq#1\Eeq{\begin{equation}#1\end{equation}}
\def\Beqo#1\Eeqo{\begin{equation*}#1\end{equation*}}
\def\Beqs#1\Eeqs{\begin{align}#1\end{align}}
\def\Beqso#1\Eeqso{\begin{align*}#1\end{align*}}

\usepackage{amsmath,amssymb,amsfonts}
\usepackage{algorithmic}
\usepackage{algorithm}
\usepackage{array}
\usepackage[caption=false,font=normalsize,labelfont=sf,textfont=sf]{subfig}
\usepackage{textcomp}
\usepackage{stfloats}
\usepackage{url}
\usepackage{verbatim}
\usepackage{multicol}
\usepackage{multirow}

\begin{document}

\begin{frontmatter}


\title{On-line reinforcement learning for optimization \\ of real-life energy trading strategy}
\author[PW,IDEAS]{Łukasz Lepak\corref{cor}}
\ead{lukasz.lepak.dokt@pw.edu.pl, lukasz.lepak@ideas-ncbr.pl}
\cortext[cor]{Corresponding author}
\author[IDEAS]{Paweł Wawrzyński}
\ead{pawel.wawrzynski@ideas-ncbr.pl}

\affiliation[PW]{organization={Warsaw University of Technology},
            addressline={Plac Politechniki 1},
            postcode={00-661},
            city={Warsaw},
            country={Poland}
            }
\affiliation[IDEAS]{organization={IDEAS NCBR},
            addressline={Chmielna 69},
            postcode={00-801},
            city={Warsaw},
            country={Poland}
            }

\begin{abstract}

An increasing share of energy is produced from renewable sources by many small producers. The efficiency of those sources is volatile and, to some extent, random, exacerbating the problem of energy market balancing. In many countries, this balancing is done on the day-ahead (DA) energy markets. This paper considers automated trading on the DA energy market by a medium-sized prosumer. We model this activity as a Markov Decision Process and formalize a framework in which an applicable in real-life strategy can be optimized with off-line data. We design a trading strategy that is fed with the available environmental information that can impact future prices, including weather forecasts. We use state-of-the-art reinforcement learning (RL) algorithms to optimize this strategy. For comparison, we also synthesize simple parametric trading strategies and optimize them with an evolutionary algorithm. Results show that our RL-based strategy generates the highest market profits. 

\end{abstract}

\begin{graphicalabstract}
\includegraphics[width=\textwidth]{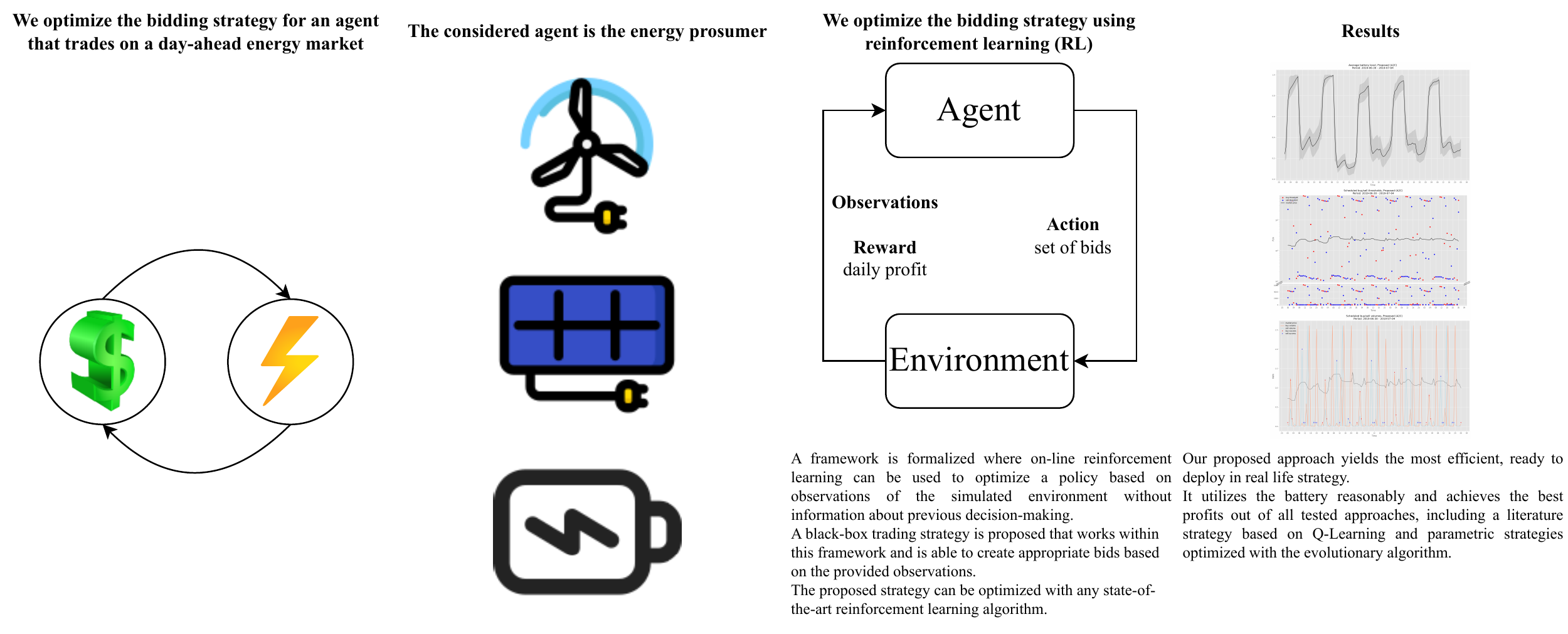}
\end{graphicalabstract}

\begin{highlights}
\item We address the problem of a prosumer who sells/buys energy on the day-ahead market.
\item We formalize this problem as a Markov Decision Process.
\item We solve the problem using state-of-the-art reinforcement learning algorithms.
\item We prove the effectiveness of our approach and compare it to other strategies.
\item We obtain efficient, ready-to-apply, automated trading strategy.
\end{highlights}


\end{frontmatter}

\section{Introduction}

In 2022, wind and solar energy represented 12\% of global electricity generation, after these shares doubled in 5 preceding years \citep{2023wiatosmotyka}. The power of wind and sunlight reaching the Earth's surface is, to some extent, random. Therefore, while the rise of renewable energy sources presents the prospect of cheap and clean energy, it also exacerbates the problem of balancing power supply and demand. 

In many countries, the main institution that balances volatile electricity supply and demand is a day-ahead energy market \cite{iria2017trading,jogunola2020consensus, prabavathi2015energy,rahimiyan2015strategic}. Every day, agents participating in this market place their buy and sell bids separately for every hour between 0 am and 11 pm the next day. Market clearing prices are then designated for each of these hours, and the bids are consequently executed or not, depending on the proposed prices. 

Here, we consider an~energy prosumer, who is an agent that (i) consumes electricity, (ii) produces electricity, and (iii) has electricity storage. What is of interest here is a~strategy for automated trading on a day-ahead energy market on behalf of this agent.

In most studies, decisions in power systems are based solely on the state of this system \cite{bose2021reinforcement,castellini2021energy,
chen2018local,
dong2021strategic,jogunola2020consensus,jogunola2021trading,
lu2019demand,may2023multi,okwuibe2022intelligent,qiu2021multi}. Moreover, the strategy for decision-making is optimized based on a~model of this system's dynamics.
We argue that (i) a~useful strategy for operation in the power system needs to be fed with data on the environment and (ii) it needs to be optimized with real-life data. Firstly, reasonable temporal energy allocation must be based on the information that makes it possible to anticipate future prices, even if they are not directly predicted. Therefore, the bids need to be based on such information. Secondly, the environment that impacts the energy prices (e.g., weather conditions) has its own temporal dynamics that are hardly possible to model but can be replayed from real-life data, which is enough for strategy optimization. 

We consider automated trading on the energy market as a problem of sequential decision-making under uncertainty in an environment that is difficult to model. A~natural approach to synthesizing a~strategy for this kind of problem is reinforcement learning (RL) \cite{2018sutton+1}. However, we face the following constraints: Firstly, trial-and-error learning in the real environment is too costly to be feasible. Secondly, we consider the dynamics of the external environment too hard to model; therefore, its simulation is impossible. Thirdly, we assume no earlier trading data is available, making off-line RL \cite{2020levine+3} impossible to apply. Instead, we develop a framework in which on-line RL can be applied to synthesize a~control policy for automated trading using a time series of observations of the environment. The policy optimized within our framework can readily be applied in real life.

Based on the above line of thought, this paper contributes as follows: 
\begin{itemize} 
\item We formalize a~framework in which on-line RL can be applied to optimize a~policy based on recorded observations of the external environment without data on earlier decision-making.
\item We design a~parametric strategy of automated bidding, which is fed with available information that makes it possible to anticipate future prices. 
\item We apply a set of state-of-the-art RL algorithms to optimize the above strategy and select the best algorithm for this purpose. The resulting strategy is fitted to the data and ready to use in real life.
\end{itemize} 

\section{Problem definition} 
\label{sec:problem} 

\subsection{Day-ahead energy market} 

Details of the day-ahead (DA) energy market are here taken from the Polish market of this kind. When created in 2000, this market was modeled on existing day-ahead energy markets in Europe. It is, therefore, typical. 

Every day between 8 am and 10.30 am, an agent participating in the market places a set of bids defined by: (i) [buy or sell] indicator, (ii) price for 1 MWh [PLN], (iii) volume [number of MWh, at least 0.1 MWh], and (iv) an hour of realization [one of 24 between 0 am and 11 pm the next day]. The bids are independent. Based on the bids placed by all agents, the clearing market price for each hour is designated. A~buy bid is accepted when its price is not below the market price for its hour. A~sell bid is accepted when its price is not above the market price for its hour. At each hour of the next day, the agents that realize their sell bids inject the declared volume of electricity into the system and get the market price for it. The agents that realize their buy bids withdraw the declared volume of electricity from the system and pay the market price for it. The agents pay small fees for entering the market, annual participation in it, and their turnover.

\subsection{Prosumer} 

The agent considered here (i) consumes electricity at random but with a~given statistical profile, (ii) produces electricity with means of limited random efficiency, such as solar panels or wind turbines, (iii) has energy storage with limited capacity and efficiency (it outputs less energy than it inputs). We also assume that the prosumer is large enough to be able to participate in a DA energy market and not large enough for its bids to change the market prices. 

At every hour, the agent may consume, produce, buy, and sell some energy. The residual energy is deposited into or taken from the energy storage. If some fraction of the residuum still remains because of the storage being full or empty, this portion is given to or taken from the market operator, and the agent is charged the corresponding penalty fee. 

An example of a~prosumer considered here is a~group (or an aggregator) of households. It cannot be a~single household, though, as the minimum volume of electricity tradeable on the market is 0.1 MWh, which is too much for a~typical single household to consume or produce.

The objective of the prosumer is to maximize its profit (or minimize its costs) by issuing optimal bids on a DA market. Essentially, the agent should buy the energy when its market price is relatively low, sell it when it is relatively high, and/or keep it in storage. The agent should also avoid paying penalty fees, thus avoiding having the storage charged or discharged entirely. Note that the problem does not quintessentially change when the prosumer does not produce nor consume electricity because then it becomes a~temporal arbitrator, and its profit still non-trivially depends on the strategy of issuing the buy/sell bids. However, if the prosumer does not have the storage, then events at different times are independent of each other, and the objective degenerates to just predicting the prosumer's own production and consumption. 

\section{Related Work} 
\label{sec:related-work} 

\paragraph{Reinforcement learning for physical world deployment}

With reinforcement learning (RL) \cite{2018sutton+1}, an agent can learn to make sequential decisions under uncertainty in a dynamic environment. The result of the learning takes the form of a~reactive policy with which the agent can transform states of the environment into actions. On-line learning achieves this goal with trial-and-error interaction of the agent with the environment. Off-line RL \cite{2020levine+3, haarnoja2018soft} optimizes the policy based on records of such interaction. A~necessary prerequisite of off-line learning is that the recorded interaction is driven by a~known, randomized control policy. This study assumes that the trial-and-error interaction is impossible due to its costs, and the above records are unavailable. 

Applying on-line RL in a simulated environment and deploying the resulting control policy in real life is also possible. However, the simulator needs to be based on a~model of the environment's dynamics, and the accuracy of this model is unavoidably limited. Consequently, the optimal policy in the simulator is not optimal in reality. 
A~lot of research \cite{2019cobbe+4, 2018packer+5, 2018zhang+2} has been devoted to assessing that optimality gap. 
Robust RL \cite{2020kamalaruban+5, 2020mankowitz+7, 2017rajeswaran+3, 2019vacaro+6, 2020zhao+2, 2020zhao+3}
aims to reduce this gap by making the simulated environment more demanding or imposing additional requirements on the resulting policy to make it more careful. But the gap has never been eliminated entirely.

\paragraph{Automated trading on the electricity market.} 

Research on automated trading on the electricity market covers various approaches. Some works introduce theoretical frameworks of bidding strategies \cite{REN2023104666, lamont1997strategic, castellini2021energy}. Many authors propose various forms of parametric bidding strategies. These strategies are optimized with methods like linear programming \cite{bakirtzis2007electricity}, genetic and evolutionary algorithms \cite{wen2001strategic, 
attaviriyanupap2005new} or stochastic optimization \cite{iria2017trading, 
liu2015bidding}. However, as a more complex bidding strategy is expected and a more complex transformation of observations into bids is required, these techniques become less effective.

With the advent of electricity prosumers, energy microgrids, and flexible price-driven energy consumption, there is an increasing need for automated decision-making and control in various activities undertaken by the energy market participants. Strategies for these agents can be optimized with reinforcement learning, which has been successfully used in financial markets \citep{XU2023107148, AVRAMELOU2024121849, MAJIDI2024121245, HUANG2024121502, JING2024121373}. Various applications of RL in power systems are reviewed in \citep{jogunola2020consensus,
yang2020reinforcement,perera2021applications}. 
\citet{nanduri2007reinforcement} analyze bidding on a DA energy market as a~zero-sum stochastic game played by energy producers willing to exercise their market power and keep their generators productive. RL is used there to optimize their bidding strategy. 
\citet{vandael2015reinforcement} analyze bidding on a DA energy market from the point of view of a~flexible buyer (who charges a~fleet of electric vehicles). His strategy is optimized with RL. 
A~number of papers is devoted to peer-to-peer trading with electricity on a~local, event-driven energy market, with RL applied to optimize the behavior of such peers \citep{
chen2018indirect,chen2018local,
bose2021reinforcement,jogunola2021trading,
qiu2021multi, 
ALSOLAMI2023102466, CUI2024109753, WANG2023108885, CAO2023108796}.  
\citet{lu2019demand} use RL and neural price predictions to optimize the scheduling of home appliances of private users. The authors assume that the electricity prices are changing and are known one hour ahead.
\citet{bose2021reinforcement} analyze a~similar setting in which the users also trade energy with each other.  
\citet{qiu2021multi} optimize the user strategies in this setting with multi-agent RL. 
\citet{angelidakis2015factored} model prosumer decision-making problem as factored Markov Decision Process with discrete states and actions, and verify this approach with a value iteration algorithm.
\citet{may2023multi} optimized peer-to-peer prosumer microgrid operations with multi-agent reinforcement learning, with their method generating higher net profits than simple fixed price biddings. 
\citet{okwuibe2022intelligent} use Q-Learning and SARSA algorithms to create simple bidding strategies and test them on German real-life data. 
\citet{dong2021strategic} use RL to optimize a~strategy of bidding on a~DA energy market by a~battery energy storage system. The authors address the dynamics of that process only to a~limited extent. Firstly, the criterion of policy optimization is on-day-ahead profit instead of a~long-term profit. Secondly, no environmental information that could impact future prices is considered, e.g., weather conditions.  

\citet{dong2021strategic} considers simultaneous trading on a~DA and hour-ahead energy markets by an energy storage operator as a~Markov Decision Process. In this MDP, consecutive days are separate episodes, so the between-day dynamics of the market are not accounted for. Discrete actions define the parameters of the bids. They are not based on external observations such as weather forecasts. This paper considers the between-day dynamics, continuous parameters of the bids, and weather forecasts. These all lead to significantly better performance of our proposed strategy. 

\section{Simulated on-line reinforcement learning with recorded environmental data}
\label{sec:simualted:RL}

Let us consider a Markov Decision Process in which the state, $\state_t$, of the environment at time $t=1,2,\dots$ is a vector composed of two sub-vectors, $\state^u_t$, and $\state^c_t$. $\state^u_t$ consists of {\it uncontrollable} coordinates; it evolves according to an~unknown stationary conditional probability 
\Beq \label{state^u} 
    \state^u_{t+1} \sim P(\cdot | \state^u_t). 
\Eeq
The sub-vector $\state^c_t$ contains {\it controllable} state coordinates. They are directly determined by the actions $\ctrl_t$ taken and the uncontrollable state coordinates, that is
\Beq \label{state^c} 
    \state^c_{t+1} = f(\state^c_t, \ctrl_t, \state^u_t, \state^u_{t+1}), 
\Eeq
where $f$ is known. The uncontrollable state variables may denote some external conditions like weather parameters. The controllable state variables may denote the internal state variable of a certain engineered mechanism whose operation is thus known in detail. 

Based on a~recorded trajectory of uncontrollable states $(\state^u_t: t=1,\dots,T)$ we can designate a~strategy of selecting actions $\ctrl_t$ based on states $\state_t$ and evaluate this strategy in a~simulation with the record $(\state^u_t: t=1,\dots,T)$ replayed. This valuation will be an~unbiased estimate of the performance of this strategy deployed in reality. Furthermore, we can replay this record repeatedly and simulate episodes of on-line RL just using $f$ \eqref{state^c} to designate consecutive values of $\state^c_t$. 

Note that the above-defined division of state variables into controllable and uncontrollable is unusual. In a~typical MDP, we assume that the state changes according to 
\Beq 
    \state_{t+1} \sim P_\state(\cdot | \state_t, \ctrl_t), 
\Eeq 
where the conditional probability $P_\state$ may be quite difficult to analyze and estimate. Therefore, a~strategy of choosing actions cannot be evaluated without bias within a~simulation based on~a~model of~$P_\state$.

\section{Model}
\label{sec:model} 

\subsection{Markov Decision Process} 

In this section, we model the automated trading on a day-ahead energy market as a~Markov Decision Process (MDP) \citep{2018sutton+1}. This MDP includes the following components: 
\begin{itemize} 
\item Time, $t=1,2,\dots$. Here, time instants denote days. 
\item Actions, $\ctrl_t \in \ctrlSpace$. An action is a set of bids for the next day in the form 
\Beq
    \ord{ volume, price, type, hour },
\Eeq 
where $type \in \{\Sell, \Buy\}$, $hour\in\{0\text{ am},1\text{ am}, \dots, 11\text{ pm}\}$. 
\item Reward, $r_t \in \real$ is equal to the profit collected during the day. 
\item States of the environment, $\state_t\in\stateSpace$. A state here is a vector that encompasses all the information about the surrounding world that may influence the market prices of electricity and the volume of its production and consumption by the prosumer. Here we divide the coordinates of the state into {\it uncontrollable}, $\state^u_t$, and {\it controllable}, $\state^c_t$, $\state_t = \ord{\state^u_t, \state^c_t}$. The uncontrollable state coordinates include an~indicator of the day within the week, an~indicator of the month within the year, energy prices for the current day, and weather forecasts. There is only one controllable state coordinate: the energy storage level. The $f$ function is known because the storage level trivially results from consuming, producing, buying, and selling energy, and the efficiency of the storage. 
\end{itemize} 
The critical assumption that allows us to distinguish uncontrollable and controllable variables is the following: The prosumer is small enough not to impact the market prices. Therefore, we may simulate its bidding and determine whether the bids are executed based on the recorded market prices. If the prosumer was large enough to actually impact the market prices, then that would not be possible, at least without an elaborate model of the impact of this prosumer on the market prices. 

\subsection{Black-box strategy and its optimization with reinforcement learning} 

In general, by a~{\it strategy}, $\pi$, we understand a~probability distribution of actions, $\ctrl_t$, conditioned on states, $\state_t$: 
\Beq \label{pi} 
    \ctrl_t \sim \pi(\cdot|\state_t). 
\Eeq 
The above action $\ctrl_t$ defines a~set of 24 pairs of bids 
\Beq \label{pi:black-box} 
    \begin{split}
    & \ord{rnd(\bar v \exp(v^B_h)), \bar p \exp(y^B_h), \Buy, h} 
    \\
    & \ord{rnd(\bar v \exp(v^S_h)), \bar p \exp(y^S_h), \Sell, h},
    \end{split} 
\Eeq 
for $h=0\text{ am}, 1\text{ am}, \dots, 11\text{ pm}$. $\bar v$ is the maximum possible energy volume the prosumer can generate from wind and solar sources in an hour (constant), while $\bar p$ denotes the 28-day average median price for the given hour. The $rnd$ function rounds resulting volumes to the first decimal point, so created bids comply with the Polish day-ahead (DA) energy market regulations. The numbers $v^B_h, y^B_h, v^S_h, y^S_h$ are coordinates of the action, which is thus $96$-dimensional. The action is produced as a~sum of the output of a~zero-mean normal noise, $\xi_t$, and a~neural network, $g$, output: 
\Beq \label{pi:RL} 
    \begin{split}
    \ctrl_t = 
    \begin{bmatrix}
    v^B_{0am}  & \dots  & v^B_{11pm} \\ 
    y^B_{0am}  & \dots  & y^B_{11pm} \\ 
    v^S_{0am}  & \dots  & v^S_{11pm} \\
    y^S_{0am}  & \dots  & y^S_{11pm}
    \end{bmatrix}
    & = g^1(\state_t; \theta) + \xi_t\circ\exp(g^2(\state_t; \theta)) \\ 
    \xi_t & \sim \normal(0, I), 
    \end{split}
\Eeq
where $g^1$ and $g^2$ are two vectors produced by the $g$ network which is fed with the state $\state_t$ and parameterized by the vector $\theta$ of trained weights; ``$\circ$'' denotes the Hadamard (elementwise) product. 

The reason to introduce the noise $\xi_t$ into the bids is exploration: By taking different actions under similar circumstances, the trading agent is able to learn to tell good actions from the inferior ones in the current state. 

To optimize the strategy \eqref{pi:RL}, we may use any algorithm of on-line reinforcement learning \citep{2001sutton+2} e.g., A2C \citep{2016mnih+many}, PPO \citep{2017schulman+4} or SAC \citep{haarnoja2018soft}. A~training consists of a~sequence of simulated trials in which the trajectory of uncontrollable states $(\state^u_t: t=1,\dots,T)$ is just replayed from the data, and the corresponding trajectory of controllable states $(\state^c_t: t=1,\dots,T)$ is designated based on the uncontrollable states, the actions selected and the function $f$ \eqref{state^c}. 

\subsection{Comparative strategy gradient-free optimization}

To compare with the above black-box strategy, we design parametric bidding strategies to optimize them with gradient-free methods such as evolutionary algorithms. To this end, let us denote by $l_t\in(0,1)$ the storage level at midnight when the bids defined in action $\ctrl_t$ start to be realized. The action $\ctrl_t$ is selected at 10.30 am on a preceding day. At this moment, $l_t$ is unknown. However, it is known which of the bids placed with $\ctrl_{t-1}$ have been and will be realized. Therefore, $l_t$ can be estimated with a~reasonable accuracy. We will denote this estimate by $\est l_t$. 

\paragraph*{Timing-based strategy (Timing)} A simple strategy may be based on an~observation that the market prices are generally low between 0 am and 3 am and high between 5 pm and 8 pm. That leads to actions comprising the following eight bids: 
\Beq \label{pi:timing} 
    \begin{split} 
    & \ord{rnd((\alpha_1 - \alpha_2\est l_t)/4), +\infty, \Buy, h}, \; h\in\{0\text{ am}, 1\text{ am}, 2\text{ am}, 3\text{ am}\}  \\ 
    & \ord{rnd((\alpha_1 + \alpha_2\est l_t)/4), 0, \Sell, h}, \quad\;\; h\in\{5\text{ pm}, 6\text{ pm}, 7\text{ pm}, 8\text{ pm}\} 
    \end{split} 
\Eeq
where $rnd$ is defined the same as for the black-box strategy and $\alpha_1, \alpha_2$ are positive coefficients. The term $\pm\alpha_2\est l_t$ results from the fact that the more we have in the storage, the less we want to buy, and the more we want to sell. The prices ($+\infty$ and $0$) are defined to ensure the bids are accepted. 

\paragraph{Opportunistic strategy (Opportunistic)} Another strategy is based on the observation that the prices generally vary, and the best thing to do is to buy when the price is relatively low and sell when it is relatively high while considering the battery level and production capabilities. That leads to the strategy in which, for each hour $h$, there is a pair of bids: 
\Beq \label{pi:arbiter}
    \begin{split} 
    & \ord{rnd(\bar v\exp(\alpha_{4h+5} + \alpha_1 \est l_t)), \bar p \exp(\alpha_{4h+7} + \alpha_3\est l_t), \Buy, h} \\ 
    & \ord{rnd(\bar v\exp(\alpha_{4h+6} + \alpha_2 \est l_t)), \bar p \exp(\alpha_{4h+8} + \alpha_4\est l_t), \Sell, h}, 
    \end{split} 
\Eeq
where $\bar v$, $\bar p$ and $rnd$ are defined the same as for the black-box strategy and $\alpha_i$ are coefficients. 
Here, we try to sell/buy varied volumes of energy at different prices based on our production capabilities and battery level. These prices and volumes are to be optimized with respect to the profit this strategy yields.

Given real data $(\state^u_t: t=1,\dots,T)$, we optimize the strategies \eqref{pi:timing} and \eqref{pi:arbiter} using a~gradient-free optimization method. In this approach, we need to be able to evaluate the strategy for any given vector of parameters. Here, an evaluation is a~simulation of events over time $t=1,\dots,T$ with the real data, given strategy in use, and calculating the resulting profit.

\section{Experimental study} 
\label{sec:simulations}

This section demonstrates the effectiveness of our proposed black-box strategy optimized with reinforcement learning. We compare it to the parametric strategies optimized with the~gradient-free CMA-ES algorithm \citep{hansen2016cma} and the strategy optimized with the algorithm from \citep{dong2021strategic}, referred to as FARL. 

\subsection{Testing environment}

Our experiments are conducted using a~custom environment simulating day-ahead energy market operations
based on real-life data from the Polish market. This environment allows for customization of various market settings, such as a~bid creation time, a~scale of the bidding prosumer (defined by the number of households), or its solar and wind energy generation capabilities. The environment is based on the Gymnasium interface \citep{gym}, making it compatible with many reinforcement learning libraries, including Stable-Baselines3 \citep{stable-baselines3}, which we use as our source of RL algorithms. 

In our experiments, we use real historical data from the following sources:
\begin{itemize}
    \item energy prices -- Polish day-ahead energy market (Fixing I).
    \item weather data -- Polish Meteorology and Water Management Institute.
    \item average energy consumption -- Polish Central Statistical Office.
\end{itemize}

As there are no publicly available historical weather forecast datasets for Poland, we generate one by noising actual weather data. 
We start each day at 10 am of the previous day in the actual data. For every hour from 11 am of the previous day to 11 pm of the currently forecasted day, we generate the forecasts as follows:
\Beq
\begin{split}
     \epsilon_t & \sim \normal(0, \sigma^2 / {24}) \\
     d_t & = \sum_{i=1}^{t} \epsilon_i \\
     x_{t}^{forecast} & = x_{t}^{actual} + d_t
\end{split}
\Eeq
where $\sigma$ is an accuracy of a 24-hour forecast, $d_t$ is a deviation for index $t$ and $x_t^{actual}, x_t^{forecast}$ are actual and forecasted weather for index $t$, respectively. For cloudiness, we assume $\sigma = 2$ Oktas; for wind speed, we assume $\sigma = 1$ m/s, and for temperature we assume $\sigma = 2$ $^{\circ}C$. Here, $t = 0$ denotes 10 am of the previous day, and we are interested in $t \in [14, 37]$, i.e., next-day forecasts. Cloudiness forecasts are clipped and projected to the nearest integers, while wind speed forecasts are clipped to be at least zero.

We also test the RL agent without the weather forecast data included in the observations. We do this to check if the weather forecasts allow the agent to define better bids, as this information impacts future energy production, consumption, and prices. 

Table \ref{tab:env-settings} depicts common environment settings used in our experiments. We set the action scheduling time to match the Polish day-ahead energy market. Battery and solar panel efficiencies reflect the efficiencies of real-life batteries and solar panels. Wind energy and solar energy limits are tuned so that daily energy production in the environment averages around 1 MWh. The number of households is set to 100 to scale the simulation to represent a medium-sized prosumer, an~aggregator, or a small energy generation facility. 

\begin{table}[!h]
\centering
\begin{tabular}{l | r}
Action scheduling time & 10.30 am \\
\hline
Battery capacity & 2 MWh \\
Battery efficiency  & 85\% \\
\hline
Maximum solar energy generation & 0.4 MWh \\
Solar panel efficiency & 20\% \\
Maximum wind energy generation & 0.05 MWh \\
Maximum wind speed for which \\
\quad wind turbines are still operational & 11 m/s \\
Number of households & 100 \\
\end{tabular}
\caption{Parameters of the environment used for experiments.}
\label{tab:env-settings}
\end{table}

Energy consumption for the given hour ($E_{c}^{h}$) is calculated as follows:
\Beq \label{eq:eng_consumption} 
    E_{c}^{h} = n \cdot E_{c\_avg}^{h} \cdot \left | 1 + \rho \right |
\Eeq
where $E_{c\_avg}^{h}$ is the average energy consumption per one household for the given hour, $n$ is the number of households, and $\rho \sim \normal(0, 0.03)$ allows the resulting energy consumption to differ each day while maintaining the average value. Equation \eqref{eq:eng_consumption} is prepared to scale well with the changing number of households. 

Solar energy production for the given hour ($E_{s}^{h}$) is based on cloudiness value from the actual weather data and is calculated as follows:
\Beq \label{eq:solar_gen} 
    E_{s}^{h} = s_{max} \cdot (1 - c/8) \cdot \eta
\Eeq
where $s_{max}$ is the maximum solar energy generation, $c \in \{0, 1, ..., 7, 8\}$ is the cloudiness value in Oktas ($0$ - clear sky, $8$ - heavy overcast) taken from the weather data and $\eta$ is the solar panel efficiency.

Wind energy production for the given hour ($E_{w}^{h}$) is based on the actual wind speed value from the weather data and is calculated as follows:
\Beq \label{eq:wind_gen} 
    E_{w}^{h} = w_{max} \cdot \frac{ws}{ws_{max}} \cdot [ws \leq ws_{max}]
\Eeq
where $w_{max}$ is the maximum wind energy generation, $ws$ is the wind speed, $ws_{max}$ is the maximum wind speed for which the wind turbines are still operational, and $[ws \leq ws_{max}] = 1 \text{ when } ws \leq ws_{max} \text{, else } 0$.

During the simulation, it may turn out that the agent has to buy missing energy or sell excess energy immediately. It is being penalized for such events. Immediate buying is realized for double the current market price, and immediate selling is realized for half the current market price so that the agent has the incentive to better plan its bids instead of relying on instant buys or sells. Also, we do not include market entry and transaction fees, as they are fixed costs independent of the bidding strategy. 

\subsection{Experiments}

\paragraph{Evolutionary algorithm}

The evolutionary algorithm CMA-ES is used to optimize the strategy defined by Equations \eqref{pi:timing} and \eqref{pi:arbiter}. It utilizes data from 2016 to 2018 as the training set and data from 2019 as the testing set. 
After training, the resulting parameters (mean values) are evaluated on a single testing interval 365 days long. Table \ref{tab:cmaes-settings} presents the parameters used for the CMA-ES algorithm. 
Customized initialization for parameters $\alpha_{4h+5}, \alpha_{4h+6}$ of Opportunistic strategy as defined in Equation \eqref{pi:arbiter} prevents the initial samples from creating bids with too high volumes, which leads the strategy to the inefficient solution of creating no bids at all. 
\begin{table}[!h]
\centering
\begin{tabular}{l | r}
\multirow{2}{*}{Initial mean value ($\mu$)} & default: $\normal(0, 1)$ \\
&\eqref{pi:arbiter}, $\alpha_{4h+5}, \alpha_{4h+6}$: $\normal(-2, 1)$ \\
Initial sigma ($\sigma$) & 1 \\
\hline
Population size & automatic ($4 + \lfloor 3 \cdot ln(n) \rfloor$) \\ 
Generations & 100 \\
\end{tabular}
\caption{Parameters of the CMA-ES algorithm used for the experiments. $n$ is the number of parameters in the strategy.}
\label{tab:cmaes-settings}
\end{table}

\paragraph{Reinforcement learning}

Reinforcement learning is used here to optimize the strategy defined in \eqref{pi:black-box}. It utilizes data from 2016 to the third quarter of 2018 as the training set, data from the fourth quarter of 2018 as the validation set, and data from 2019 as the testing set. The training is done in 90 days long intervals randomly selected from the training set.  Periodically, evaluation is done on a single validation interval 90 days long. After the training timesteps budget is depleted, the model for which the highest reward on validation interval was achieved is evaluated on the single testing interval 365 days long. Parameters used for the A2C algorithm (which achieved the best results in our experiments) are presented in Table \ref{tab:a2c-settings}.

\begin{table}[!h]
\centering
\begin{tabular}{l | r}
Timesteps & 4 500 000 \\
Evaluation frequency & 9 000 \\
Episode length & 90 \\
\hline
Action space & 96, range $[-3, 3]$ \\
Observation space with weather data & 141, normalized \\
Observation space without weather data & 69, normalized \\
Reward space & $(-\infty, \infty)$ \\
\hline
Learning rate (\textit{learning\_rate}) & 0.0001 \\
Number of update steps (\textit{n\_steps}) & 90 \\
Discount (\textit{gamma}) & 0.9 \\
GAE coefficient (\textit{gae\_lambda}) & 0.9 \\
Entropy coefficient (\textit{ent\_coef}) & 0.0 \\
Value function coefficient (\textit{vf\_coef}) & 0.5 \\
RMSprop as optimizer (\textit{use\_rms\_prop}) & True \\
RMSprop epsilon (\textit{rms\_prop\_eps}) & 0.00001 \\
Use gSDE (\textit{use\_sde}) & False \\
\hline
Hidden layers neurons (\textit{net\_arch}) & 200 \\
Log standard deviation initial value (\textit{log\_std\_init}) & -1 \\
Normalize input (\textit{normalize\_images}) & False \\
Activation function (\textit{activation\_fn}) & tanh \\
Orthogonal initialization (\textit{ortho\_init}) & True
\end{tabular}
\caption{Parameters of the A2C algorithm used for experiments. Names in brackets are taken from the Stable-Baselines3 library \citep{stable-baselines3}. Parameters not present in this table use default values. Neural network architecture is the same for actor and critic networks.}
\label{tab:a2c-settings}
\end{table}

The action space is limited to $[-3, 3]$, allowing the agent to define prices and volumes up to $e^3\approx20$ times smaller/larger than the 28-day median hour price and maximum possible hourly energy generation volume. 

The observation of the environment's state (141 values) is passed to the agent at bid placing time and contains the following information:
\begin{itemize}
    \item prices of energy at the current day for every hour (24 values) -- these prices result from the bids created the day before.
    \item average energy consumption for $n$ households for every hour, $n \cdot E_{c\_avg}^{h}$ (24 values) -- this is statistical data about consumption, the actual consumption data is designated according to \eqref{eq:eng_consumption}.
    \item current relative battery charge (1 value).
    \item estimated relative battery charge at midnight (1 value).
    \item one-hot encoded information about the current month (12 values).
    \item one-hot encoded information about the current day of the week (7 values).
    \item cloudiness, wind speed, and temperature forecasts for each hour of the next day (72 values).
\end{itemize}

For comparison, we also applied the FARL algorithm from \citep{dong2021strategic}, which is a~conceptually different approach to optimize a black-box bidding strategy. We fed it with the same training, evaluation, and test data discussed above.
The FARL algorithm optimizes discrete actions. Thus, in order to use it in our environment, we perform action discretization in the following way ($b$ is equal to maximum battery capacity, $\bar p$ is defined the same as for the black-box strategy):
\begin{itemize}
    \item eleven capacity levels ($0, \frac{b}{10}, \frac{2b}{10}, ..., \frac{8b}{10}, \frac{9b}{10}, b$), \\ eleven price levels ($0, \frac{\bar p}{5}, \frac{2\bar p}{5}, ..., \frac{8\bar p}{5}, \frac{9\bar p}{5}, +\inf$), \\ separate bids for buying and selling, which gives $11 \cdot 11 \cdot 2 = 242$ different actions. Note that many actions refer to doing nothing, e.g., buying/selling zero capacity for different prices.
\end{itemize}
We used observations matching the original paper, which are:
\begin{itemize}
    \item price from the same hour of the previous day ($p_p$, normalized to range $[-1, 1]$)
    \item action from the previous hour ($a_{t-1}$, volume normalized to range $[-1, 1]$ (+ for buy bid, - for sell bid), price normalized to range $[-1, 1]$)
    \item relative battery state estimate at 0 am of the trading day ($b_t$, range $[0, 1]$)
    \item current hour ($t$, provided as a one-hot vector with 1 on the current hour index ($0-23$), else 0)
\end{itemize}
Note that in order to produce bids for each hour of the trading day, this algorithm is run in a~24-step episode. Also, this algorithm does not consider any external information, such as weather forecasts.
We have also prepared a wrapper for our original environment, which converts actions and observations and allows the FARL algorithm to be executed with timesteps representing one hour instead of one day.

Parameters of the FARL algorithm are presented in Table~\ref{tab:farl-settings}. The discount is set to 1 to match it with the original paper. The total number of timesteps is set to match the number of days seen throughout the training, as the environment for FARL uses hours as timesteps instead of days. 

\begin{table}[!h]
\centering
\begin{tabular}{l | r}
Timesteps & 108 000 000 \\
Evaluation frequency & 2 400 \\
Episode length & 24 \\
\hline
Action space & 242, discrete \\
Observation space & 28, normalized \\
Reward space & $(-\infty, \infty)$ \\
\hline
Learning rates ($\alpha$, $\beta$) & 0.0001 \\
Discount ($\gamma$) & 1.0 \\
\hline
\multirow{2}{*}{Exploration rate ($\varepsilon$)} & 1.0 - 0.1, linear decrease during \\
&10\% of timesteps, later constant \\
\end{tabular}
\caption{Parameters of the FARL algorithm used for experiments. }
\label{tab:farl-settings}
\end{table}

\begin{table}[!h]
\centering
\begin{tabular}{l | r}
Strategy & Achieved income \\
\hline
Reference & 46214.33 $\pm$ 16.84 \\
\textbf{Proposed (A2C)} & \textbf{60027.13 $\pm$ 398.11} \\
Proposed (A2C, no weather forecasts) & 56006.59 $\pm$ 321.92 \\
FARL & 29820.81 $\pm$ 4638.89 \\
Timing (CMA-ES) & 37178.98 $\pm$ 1029.76 \\
Opportunistic (CMA-ES) & 43898.02 $\pm$ 2227.74 \\
\end{tabular}
\caption{Final balances achieved on testing data by different strategies averaged over five testing runs. Descriptions in brackets denote additional information about the given strategy. }
\label{tab:balances}
\end{table}

\begin{table}[!h]
\centering
\begin{tabular}{l | r}
Algorithm used & Achieved income \\
\hline
\textbf{A2C} & \textbf{60027.13 $\pm$ 398.11} \\
A2C (GAE coefficient set to 0) & 5842.54 $\pm$ 6832.48 \\
PPO & 42234.52 $\pm$ 5864.02 \\
SAC & -1447.54 $\pm$ 12964.64 \\
\end{tabular}
\caption{Final balances achieved on testing data by the proposed strategy optimized with different reinforcement learning algorithms averaged over five testing runs.}
\label{tab:rl_alg_balances}
\end{table}

\begin{figure*}[!htb]
    \includegraphics[width=\textwidth]{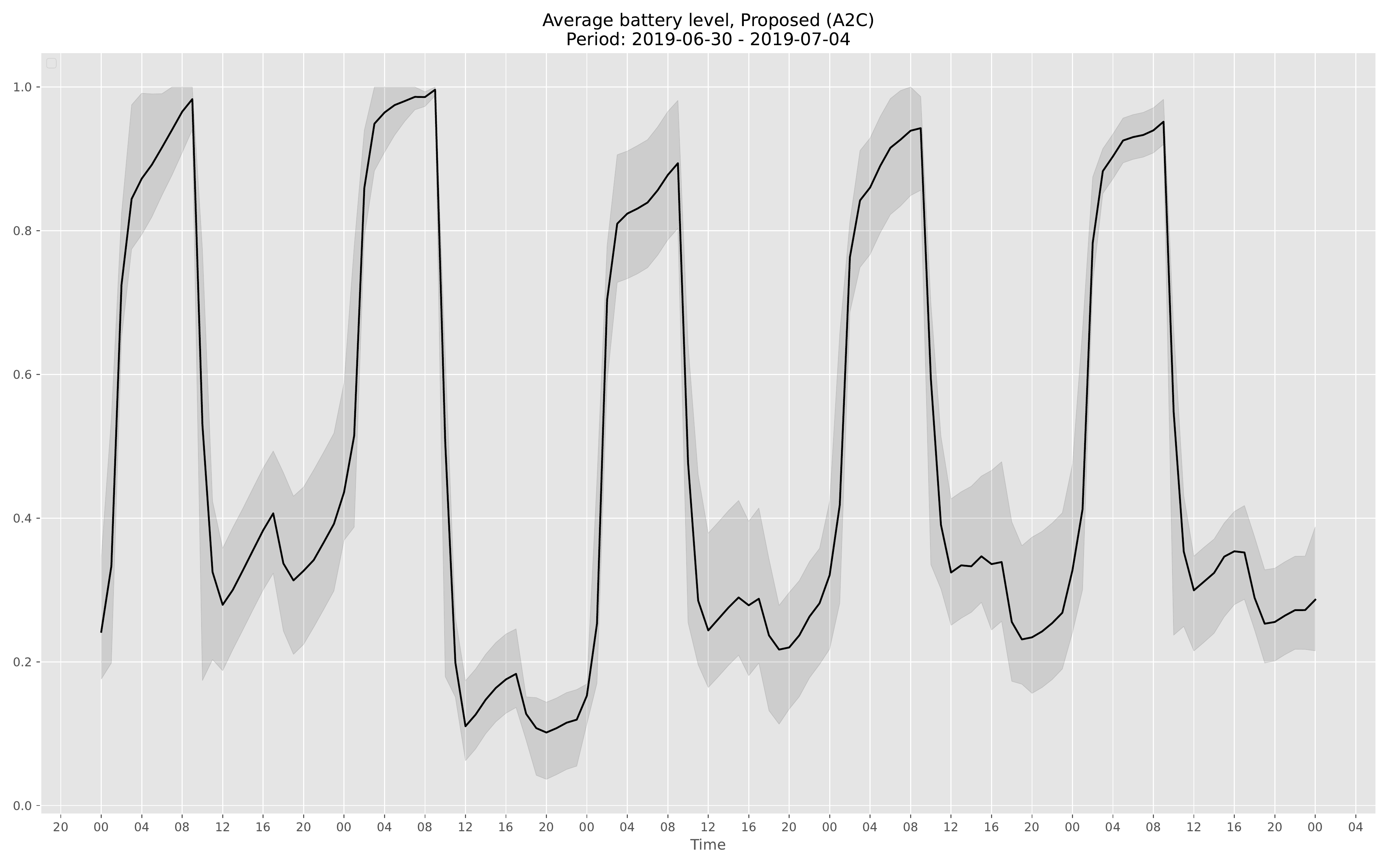}
    \caption{Battery levels for the proposed strategy during five simulation days averaged over five testing runs. }
    \label{fig:a2c_battery}
\end{figure*}

\begin{figure*}[!htb]
    \includegraphics[width=\textwidth]{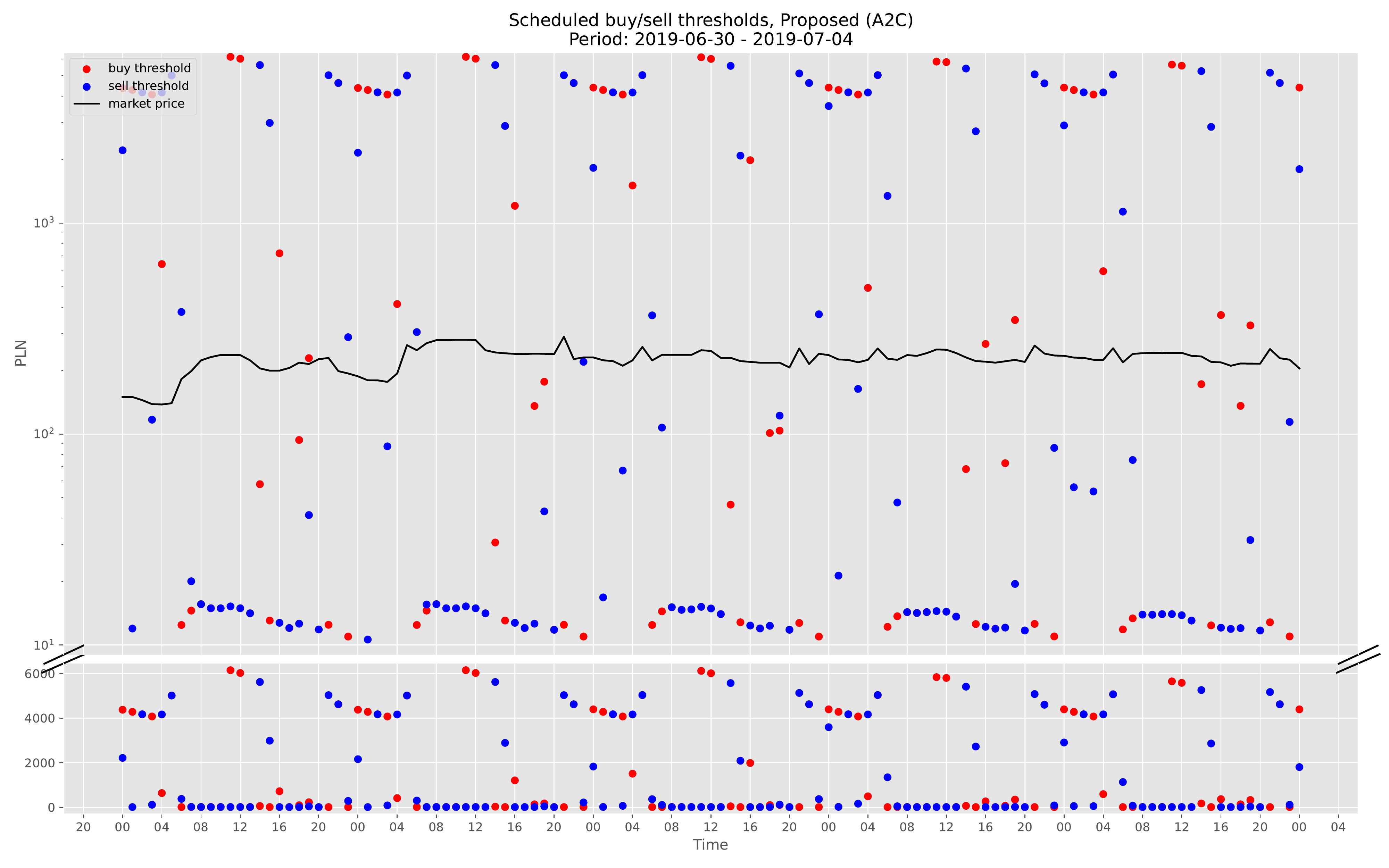}
    \caption{Bids prices for the proposed strategy during five simulation days taken from the best testing run. The vertical axis of the upper plot is in a logarithmic scale. }
    \label{fig:a2c_prices}
\end{figure*}

\begin{figure*}[!htb]
    \includegraphics[width=\textwidth]{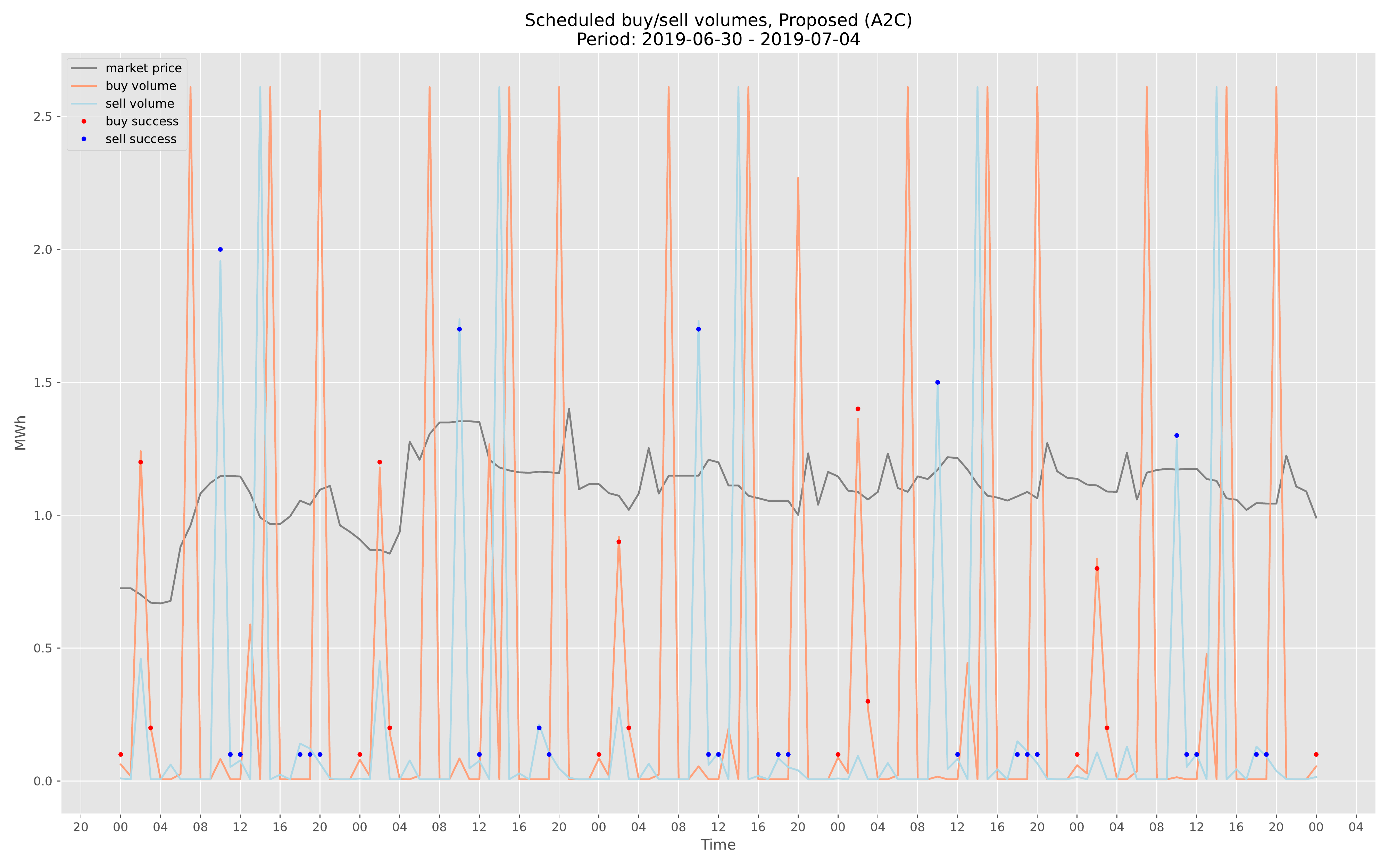}
    \caption{Bids amounts for the proposed strategy during five simulation days taken from the best testing run. }
    \label{fig:a2c_amounts}
\end{figure*}

\begin{figure*}[!htb]
    \includegraphics[width=\textwidth]{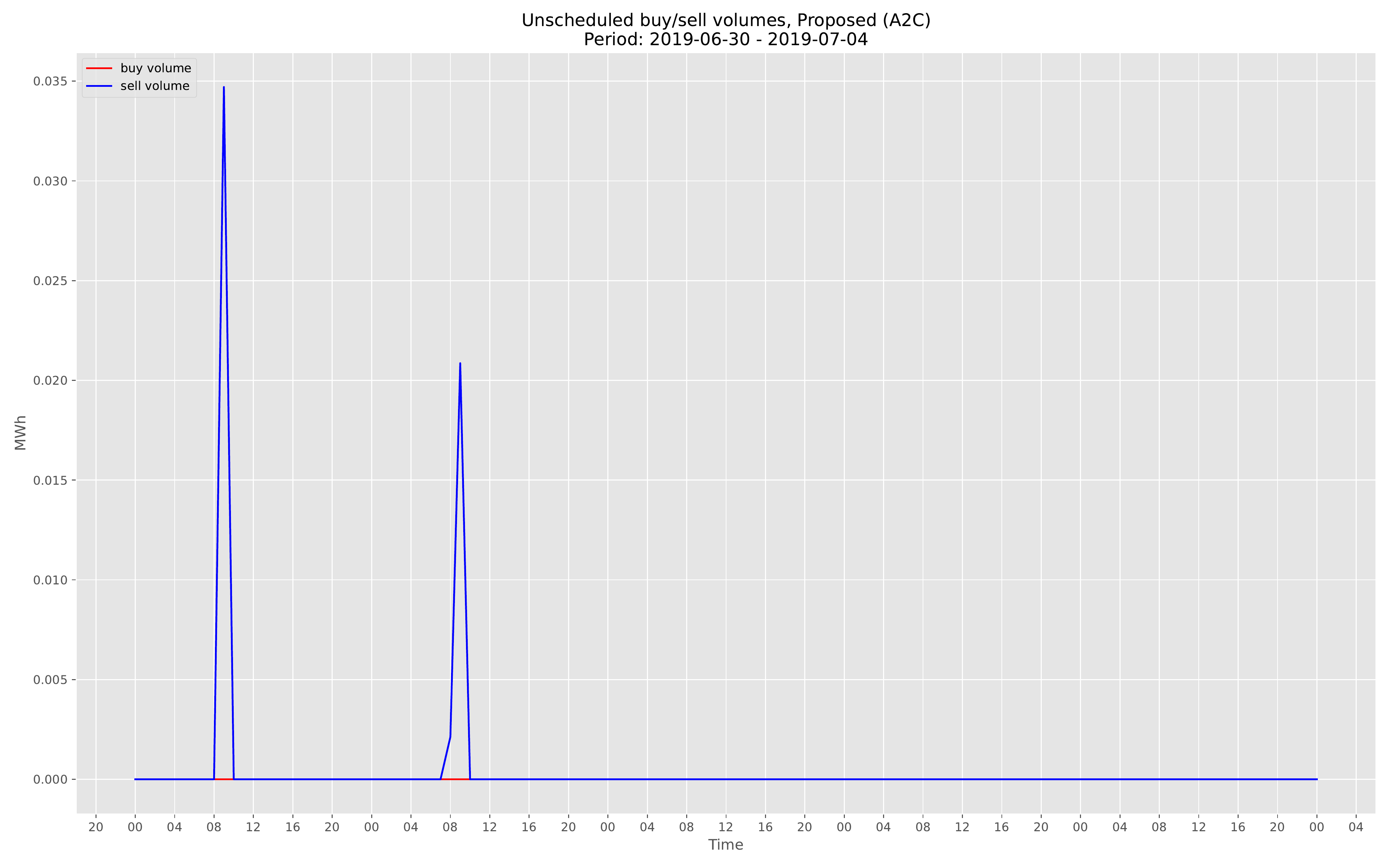}
    \caption{Unscheduled transactions' amounts for the proposed strategy during five simulation days taken from the best testing run. }
    \label{fig:a2c_unscheduled}
\end{figure*}

\subsection{Results}

Results of experiments are shown in Tables \ref{tab:balances}-\ref{tab:a2c-diff-batteries} and Figures \ref{fig:a2c_battery}-\ref{fig:a2c_unscheduled}.
We define the reference balance of the given day as the difference between the energy produced and consumed multiplied by this day's average energy price. We then calculate the sum of these reference balances during the whole simulation. Note that it is moderately difficult to achieve the reference balance. The agent mostly consumes the energy in the evenings, when it is expensive, and produces it when its price is average. Therefore, to reach the reference balance, the agent should manage the uncertainty of the prices and not buy energy at hours when it needs it.

Table \ref{tab:balances} presents average final balances from test simulations, averaged over five test runs on different random seeds. We can see that the proposed black-box strategies optimized with the A2C algorithm achieve the best return on the test simulation, beating the reference balance and the strategies optimized with the CMA-ES and FARL algorithms. Also, the A2C-trained strategy utilizing weather forecasts as part of its observations achieves higher returns than the A2C-trained strategy without those observations. 

In Figures \ref{fig:a2c_battery}-\ref{fig:a2c_unscheduled}, we look into five days from the middle of the testing simulations. For battery levels plots, we show the average relative battery charge with streaks around graphs indicating the range of levels from all testing simulations, while the other plots were taken from the testing run that achieved the best balance. There is an unscaled market price graph on the bid volumes plot, which allows for easy identification of whether the successful bid was realized when the price was high or low.

It is seen in Figure \ref{fig:a2c_battery} that the strategy trained with RL behaves reasonably: It charges the battery at about 0 am when the prices are low and discharges at about 8 am when they are high. Unscheduled buys/sells, which are costly, are rare, as seen in Figure \ref{fig:a2c_unscheduled}. 

In Figures \ref{fig:a2c_prices} and \ref{fig:a2c_amounts}, it can be seen that the proposed strategy places very high buy prices or very low sell prices with non-zero volume, whether it wants the bid to be executed. When the bid is not to be executed, its price is either set very high for sell bids and very low for low bids, or the corresponding volume is set to zero. 

The simple Timing strategy behaves reasonably, as it just places bids at times with the lowest and highest energy prices for buy and sell bids, respectively, which allows it to achieve profits. However, due to its simplicity, it cannot represent sufficiently complex behavior to respond efficiently to diverse circumstances, which makes it achieve worse results than the reference.

The more elaborate Opportunistic strategy achieves a~better average return than its simple Timing counterpart. However, it is still unable to improve the results over the reference balance. Also, the range of returns of this strategy is larger than those of other strategies. This strategy usually tries to place selling bids with low prices and high volumes and buying bids with high prices and low volumes. The wide range of returns between runs indicates that the optimization of this strategy is susceptible to getting stuck in local optima. Therefore, it is unlikely that this strategy's globally optimal parameters have been found with the CMA-ES algorithm. 

The bidding strategy developed by the FARL algorithm barely exhibited reasonable behavior. This algorithm is based on {\it Q-Learning} with function approximation applied in a~rather non-standard way: It learns to make a~sequence of $24$ bids, each time only having access to the previous bidding and several other variables. Reasonable bidding based on that information was not possible. 

Table \ref{tab:rl_alg_balances} shows average final balances from test simulations for the proposed black-box strategy optimized with different RL algorithms. A2C achi\-eves the best and most stable results out of all tested algorithms. PPO is, on average, below the reference. The worst of all tested algorithms is SAC, which profit averages around 0 with very unstable behavior -- some runs generate some profit, while others end up with significant losses. 

Also, in Table \ref{tab:rl_alg_balances}, we show the result of the proposed strategy trained with the A2C algorithm but with the Generalized Advantage Estimate (GAE) coefficient set to 0 to match the SAC approach to estimating expected future returns. SAC and A2C with the GAE coefficient set to 0 do not exhibit reasonable behavior and cannot achieve profits close to the best A2C strategy or the reference. We conclude that ceasing to use the eligibility traces (setting the GAE coefficient to 0) makes RL fail in this environment.

\subsection{Battery capacity optimization} 

The battery is often the largest part of the prosumer installation cost. Our proposed approach can be readily used to choose the battery from the possible options. It is enough to optimize each option's strategy and compare the incomes with the battery costs. 

Table \ref{tab:a2c-diff-batteries} presents the income gained with the strategy optimized with the A2C algorithm with weather forecasts as input, depending on the battery capacity. 
It is seen that the larger the battery capacity, the more the prosumer buys cheap and sells high, thus the larger the income. However, the prosumer mostly earns from what he produces.

\begin{table}[!h]
\centering
\begin{tabular}{l | r}
Maximum battery capacity (MWh) & Achieved income \\
\hline
1.0 & 41436.84 $\pm$ 1721.37 \\
1.5 & 50643.02 $\pm$ 1214.04 \\
\textbf{2.0} & \textbf{60027.13 $\pm$ 398.11} \\
\end{tabular}
\caption{Final balances achieved on testing data for different maximum battery capacities with the proposed strategy optimized with the A2C algorithm averaged over five testing runs.}
\label{tab:a2c-diff-batteries}
\end{table}

\subsection{Discussion} 

The optimal bidding strategy among those analyzed here is based on neural networks trained with reinforcement learning and fed with weather forecasts. Weather impacts the production of energy (e.g., by wind turbines), its consumption (e.g., by air conditioners), and thus its prices. Consequently, optimal bids need to be based on these forecasts. We have tried several RL algorithms. A2C yielded the best performance. PPO achieved slightly worse results, which may be due to instability periods during training. The algorithm that was especially disappointing was SAC. This algorithm is based on the action-value function with the action (bid parameters in this case) having 96 dimensions. Under these circumstances, the action-value function was impossible to approximate with sufficient accuracy, hence poor performance. Also, A2C with the GAE coefficient set to 0 showed poor performance. Based on this, we think it is difficult to make accurate future return approximations to use instead of actual returns, and because of that, algorithms that rely on these approximations (i.e., SAC) fail in the considered environment. 

Timing parametric strategy with parameters optimized with the CMA-ES evolutionary algorithm behaved worse than strategies based on neural networks. 
One can come up with even more elaborate strategies than \eqref{pi:timing}, but then this strategy will have more parameters, and their optimal values will be more difficult to find for any gradient-free optimization algorithm, as the search space will be of higher dimensionality. This can be seen with the Opportunistic strategy defined in \eqref{pi:arbiter}.

The bidding strategy learned with the FARL algorithm delivered disappointing results, even worse than those achieved with optimized parametric strategies. Its management of available information proved insufficient to map available observations into actions effectively.

\section{Conclusions} 
\label{sec:conclusions} 

In this paper, we proposed a framework for optimization of bidding strategy on a~day-ahead energy market based on simulations and real-life data. 
We have used state-of-the-art reinforcement learning to optimize two strategies that produced bids for this market. One of them was fed with weather forecasts, and the other was not. The strategy fed with weather forecasts produced the highest financial return out of all tested approaches. It is readily applicable in the market.

\section*{Data statement}
The data and code used for the experiments will be made available on request.

\section*{Funding}
This research did not receive any specific grant from funding agencies in the public, commercial, or not-for-profit sectors.

\section*{Declaration of competing interest}
The authors declare that they have no known competing financial interests or personal relationships that could have appeared to influence the work reported in this paper.

\section*{Author contributions}
\textbf{Łukasz Lepak}: Conceptualization, Data curation, Formal analysis, Investigation, Methodology, Project administration, Resources, Software, Validation, Visualization, Writing - original draft, Writing - review \& editing.

\textbf{Paweł Wawrzyński}: Conceptualization, Formal analysis, Investigation, Methodology, Project administration, Resources, Supervision, Validation, Writing - original draft, Writing - review \& editing.

\bibliographystyle{elsarticle-num}

\end{document}